\documentclass[letterpaper, 10 pt, conference]{ieeeconf}  

\usepackage{CJKutf8}
\usepackage{cite}
\usepackage{bm}
\usepackage{makecell}
\usepackage{booktabs}
\usepackage{multirow}
\usepackage{amsmath} 
\usepackage{amssymb}  
\usepackage{amsfonts}
\usepackage{ multicol} 

\usepackage{graphics} 
\usepackage{graphicx}
\usepackage{epsfig} 
\usepackage{mathptmx} 
\usepackage{times} 
\usepackage{subcaption} 
\usepackage{tikz}
\usetikzlibrary{calc}

\usepackage{mathrsfs}
\usepackage{indentfirst}
\usepackage{multirow}
\usepackage{adjustbox}

\usepackage{color}      
\usepackage{pifont}     
\usepackage{algorithmic}
\usepackage{graphicx}
\usepackage{textcomp}
\usepackage{bbding}

\usepackage[table,dvipsnames,xcdraw]{xcolor} 

\usepackage{colortbl} 
\usepackage{diagbox} 
\usepackage{siunitx}

\usepackage[colorlinks=true,linkcolor=blue,citecolor=green,urlcolor=blue,]{hyperref}

\usepackage{algorithm}
\usepackage{algorithmic}

\usepackage{subcaption}

\colorlet{colorFst}{Green!25}       
\colorlet{colorSnd}{SpringGreen!45} 
\colorlet{colorTrd}{Yellow!30}      
\colorlet{colorLow}{darkgray!30}    
\definecolor{R1}{HTML}{E97451}
\definecolor{R2}{HTML}{008080}
\definecolor{R3}{HTML}{0047AB}
\colorlet{cmt}{darkgray!80}    
\colorlet{supp}{darkgray!50}    

\newcommand{\blackx}{{\color{black}\ding{55}}}

\newcommand{\fs}{\cellcolor{colorFst}}   
\newcommand{\nd}{\cellcolor{colorSnd}}      
\newcommand{\rd}{\cellcolor{colorTrd}}      

\IEEEoverridecommandlockouts                              

\usepackage{xcolor}
\definecolor{NIDARcolor}{RGB}{0, 128, 128} 

\author{
Junjie Zhang$^{1}$, Jie Yin$^{2}$, Kefei Qian$^{1}$, Jie Li$^{1}$, Mengpei Jia$^{3}$*, Yajuan Dun$^{3}$, Wenbo Chu$^{4}$, Guofa Li$^{1}$\\[0.5em]
{Project Page:
\href{https://nidar-web.github.io}{https://nidar-web.github.io}}
\thanks{This work was supported by the National Key Research and Development Program of China under Grant 2025YFB2606504.}%
\thanks{$^{1}$Chongqing University, $^{2}$Shanghai Jiao Tong University, $^{3}$Ministry of Industry and Information Technology,
$^{4}$Western China Science City Innovation Center of Intelligent and Connected Vehicles, $^*$ Corresponding author}%
}

\definecolor{nircolor}{RGB}{0,100,150} 

\definecolor{darcolor}{RGB}{0,100,150}    

\title{\LARGE \bf
\textcolor{nircolor}{NI}\textcolor{darcolor}{DA}\textcolor{nircolor}{R}: NIR-Guided Intrinsic Decomposition for Scalable Scene-Agnostic LiDAR Intensity Reconstruction

}

\begin{document}

\maketitle
\thispagestyle{empty}
\pagestyle{empty}

\begin{abstract}

LiDAR return intensity provides complementary surface-response cues for robotic perception and state estimation, yet many simulation pipelines omit it or reproduce it using reconstruction methods that require real intensity supervision and per-scene optimization. These requirements increase data-collection and fitting costs and limit reuse across simulated scenes. We present \href{https://nidar-web.github.io/}{NIDAR}, a feed-forward framework that synthesizes dense intensity-like observations from RGB appearance and simulator geometry. NIDAR combines pretrained pseudo-NIR translation, hierarchical intrinsic decomposition, geometry-aware modulation, and source-domain distribution calibration to transfer reflectance-related image cues to simulated point clouds. Its learned components are trained offline using Waymo data; their weights and calibration remain fixed during evaluation on Waymo and nuScenes. Deployment therefore requires neither target-scene intensity labels nor target-scene gradient-based fitting. The reported comparisons show competitive pixel-wise accuracy and favorable structural and perceptual fidelity against the evaluated reconstruction baselines. A controlled pseudo-NIR-versus-RGB diagnostic further shows that the pseudo-NIR prior is most beneficial when used through the paper-aligned reflectance-and-remapping route, rather than as a simple direct intensity regressor. We further integrate NIDAR with Unreal Engine 5, Isaac Sim, and a generative LiDAR pipeline. Two intensity-aware SLAM systems evaluated in two simulated indoor scenes suggest potential downstream utility, but do not constitute real-robot validation. NIDAR therefore offers a scalable intensity-synthesis interface for the evaluated settings; cross-wavelength, camera-configuration, embedded, and real-sensor validation remain future work.

\end{abstract}


\section{INTRODUCTION}
LiDAR (Light Detection and Ranging) has become a core sensor in autonomous driving~\cite{cadena2016past}, robotics~\cite{yin2024ground,yin2021m2dgr}, and 3D scene reconstruction~\cite{zhang2025towards}, due to its ability to provide accurate 3D geometry with strong robustness to illumination variations. Beyond spatial coordinates, LiDAR also measures intensity, which reflects surface response under Near-Infrared (NIR) illumination and encodes complementary material and physical cues.
Prior studies have shown that intensity can benefit 3D object detection~\cite{guo2026lidar}, semantic segmentation~\cite{viswanath2025reflectivity}, and SLAM~\cite{wang2026pizza,wang2021intensity,tian2026ultra}. These findings motivate more faithful intensity modeling for robotic perception and simulation.

Despite rapid progress in LiDAR simulation and novel view synthesis, most existing methods prioritize geometric reconstruction while giving less attention to intensity modeling. Graphics engine-based simulators~\cite{dosovitskiy2017carla,koenig2004gazebo} commonly approximate intensity using ray tracing or handcrafted reflectance parameters under simplified material and sensor assumptions. Neural reconstruction approaches based on NeRF~\cite{mildenhall2021nerf} or 3D Gaussian rendering~\cite{kerbl2023gaussian} can incorporate intensity into LiDAR re-simulation~\cite{zhang2023nerflidar,huang2023nfl}, but typically require real intensity supervision and per-scene optimization. These requirements limit their scalability when many scenes must be simulated. This motivates a feed-forward intensity-synthesis interface that can be reused across the evaluated scenes without scene-specific network fitting.

\begin{figure*}[htbp]
\centering
  \includegraphics[width=0.85\textwidth]{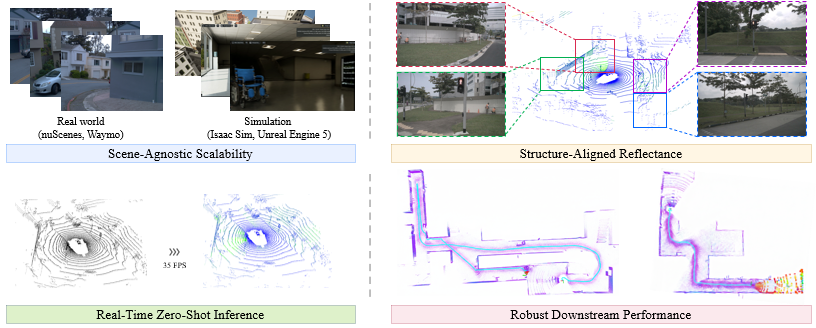}
  \caption{NIDAR generates structure-aligned intensity-like observations without per-scene network optimization in the tested settings, and is evaluated on two datasets, three simulation pipelines, and two simulated SLAM cases.}
  \label{fig:teaser}
\vspace{-0.2cm}
\end{figure*} 

To address this gap, we propose \textbf{NIDAR}, an NIR-guided intrinsic-decomposition framework for scalable LiDAR intensity synthesis. Our working hypothesis is that visible-spectrum appearance contains reflectance-related cues that are informative of, although not identical to, material response in the NIR band. NIDAR first synthesizes a pseudo-NIR image from RGB input, decomposes it into reflectance-related and shading components, and then maps the learned representation to 3D points using geometry-aware modulation and distribution calibration. The output is an intensity-like signal rather than a sensor-independent reconstruction of absolute returned power. Here, ``scene-agnostic'' denotes reuse without scene-specific network fitting in the evaluated settings; it does not imply calibration-free transfer or sensor-independent radiometric reconstruction.

We implement the intrinsic stage using a hierarchical UNet+DeepLabV3 design. After offline training and source-domain calibration, its network weights remain fixed during evaluation; unlike reconstruction-based baselines, it does not perform gradient-based fitting for each evaluated scene. Accordingly, our contribution lies primarily in the system formulation and robotics-oriented evaluation: we organize complementary spectral, intrinsic, geometric, and calibration components into a reusable intensity interface and examine its behavior across datasets and simulation pipelines. The main contributions are:

\begin{itemize}
    \item We formulate dense LiDAR intensity synthesis as a reusable simulation interface that converts RGB appearance and simulator geometry into intensity-like observations without per-scene network optimization in the evaluated settings.
    
    \item We develop a structured pipeline that combines pretrained spectral translation, hierarchical intrinsic decomposition, geometry-aware modulation, and source-domain distribution calibration, while explicitly distinguishing a reflectance-related latent representation from calibrated LiDAR return intensity.
    
    \item We evaluate NIDAR on Waymo and nuScenes, add a controlled pseudo-NIR-versus-RGB diagnostic showing that the spectral prior primarily benefits the reflectance-calibration route, integrate it with Unreal Engine 5, Isaac Sim, and LiDAR-Diffusion, and assess its potential downstream utility using two intensity-aware SLAM systems in two simulated indoor scenes.
\end{itemize}

\section{Related work}

\subsection{Graphics Engine-Based LiDAR Simulation}

Early LiDAR simulation systems are primarily built upon graphics engines, including CARLA\cite{dosovitskiy2017carla}, Webots\cite{michel2020webots}, CoppeliaSim\cite{rohmer2013vrep}, Gazebo\cite{koenig2004gazebo}, and Isaac Sim~\cite{nvidia2022isaacsim}. These platforms simulate LiDAR scanning via ray casting, producing 3D coordinates and range measurements, and can partially model noise and raydrop effects. Random raydrop typically arises from sensor noise, whereas physical raydrop is caused by low-reflectivity materials such as glass. To enhance realism, several works introduce learned echo models or neural modules to better approximate point cloud distributions. For instance, Isaac Sim leverages RTX ray tracing to simulate raydrop on transparent objects~\cite{nvidia2022isaacsim}.

Despite their flexibility and efficient deployment, engine-based simulators mainly focus on geometric fidelity and ray behavior, while the intensity channel is either omitted or heuristically approximated.

\subsection{LiDAR Reconstruction from Real-world Data}

Engine-based simulation relies on manually constructed CAD scenes, which is labor-intensive and suffers from a notable domain gap to real-world data. To alleviate this issue, reconstruction-based approaches from real scans have been explored. SqueezeSegV2~\cite{wu2018squeezesegv2} leverages unsupervised domain adaptation to bridge synthetic-to-real discrepancies, while Fang et al.~\cite{fang2020augmented} propose a hybrid framework that fuses real background scans with synthetic foreground objects using a physically-aware LiDAR renderer. PCGen and LiDARsim~\cite{2022PCGenPC,2020LiDARsimRL} further learn raydrop characteristics via FPA and U-Net, respectively, reducing the sim-to-real gap.

With the emergence of Neural Radiance Fields (NeRF)~\cite{qu2024implicit} and 3D Gaussian Splatting~\cite{kerbl2023gaussian}, methods such as NeRF-LiDAR and UniSim~\cite{zhang2023nerflidar,yang2023unisim} introduce neural implicit representations for LiDAR re-simulation. UniSim exploits multimodal cues for improved reconstruction, with later extensions supporting dynamic scenes. To improve efficiency, LiDAR-RT~\cite{2025lidarrt} combines 3D Gaussians with hardware ray tracing for real-time re-simulation and intensity rendering.
Despite these advances, many reconstruction-based methods rely on real intensity measurements for supervision or per-scene optimization. This dependence limits their scalability beyond the captured scenes.

\subsection{LiDAR Simulation Based on Data Generation}

To decouple simulation from explicit physical modeling and sensor constraints, generative approaches directly synthesize LiDAR data. Methods based on GANs\cite{goodfellow2014generative}, VAEs\cite{kingma2013vae}, and diffusion models~\cite{ho2020denoising} have been proposed, including PointFlow~\cite{yang2019pointflow}, GECCO~\cite{tyszkiewicz2023gecco}, LiDAR-generation~\cite{2018DeepGM}, LiDAR-Diffusion~\cite{ran2024towards}, and R2DM~\cite{nakashima2024r2dm}. In particular, LiDAR-generation and LiDAR-Diffusion employ diffusion models to simulate raydrop and enable controllable LiDAR generation.
While generative models offer strong diversity and scalability, their intensity channel is often treated as another learned feature dimension rather than being explicitly decomposed into reflectance-, geometry-, and sensor-related factors.

Table~\ref{sample_tab} summarizes the comparison. NIDAR uses RGB appearance together with available simulator or reconstruction geometry, and it does not optimize network parameters separately for each evaluated scene. Its learned components and calibration are nevertheless trained offline using source-domain LiDAR intensity.

\begin{table}[t]
    \centering
    \caption{Comparison of LiDAR intensity simulation methods.}
    \renewcommand{\arraystretch}{1.15}
    \label{sample_tab}
    \resizebox{\columnwidth}{!}{%
    \begin{tabular}{lcccc}
        \hline
        {Method} & {Paradigm} & \makecell[c]{{Intensity}\\{channel}} & \makecell[c]{\textbf{Per-scene}\\{fitting}} & \makecell[c]{{Scene-specific}\\{intensity sup.}} \\
        \hline
        \makecell[l]{Webots\cite{michel2020webots}, Gazebo, UE5\cite{koenig2004gazebo}} & Graphics & No & No & No \\
        \makecell[l]{Isaac Sim (built-in)\cite{nvidia2022isaacsim}} & Graphics & Yes & No & No \\
        \makecell[l]{PCGen\cite{2022PCGenPC}, LiDARsim\cite{2020LiDARsimRL},LiDAR-4D\cite{zheng2024lidar4d} \\LiDAR-NeRF\cite{tao2023lidarnerf},LiDAR-RT\cite{2025lidarrt}} & Recon. & Yes & Yes & Yes \\
        LiDAR-generation\cite{ran2024towards}, LiDAR-Diffusion\cite{2018DeepGM} & Gen. & No & No & No \\
        \textbf{NIDAR (Ours)} & \fs\textbf{Multiple} & \fs\textbf{Yes} & \fs\textbf{No} & \fs\textbf{No} \\
        \hline
    \end{tabular}%
    }

    \raggedright
    \footnotesize
    Here, per-scene fitting denotes learning-based optimization on a target scene (e.g., NeRF/GS), rather than manual authoring. Scene-specific intensity supervision excludes offline source-domain supervision used to train NIDAR.
    
    \vspace{-3mm}
\end{table}

\begin{figure*}[htbp]
\centering
\includegraphics[width=0.85\linewidth]{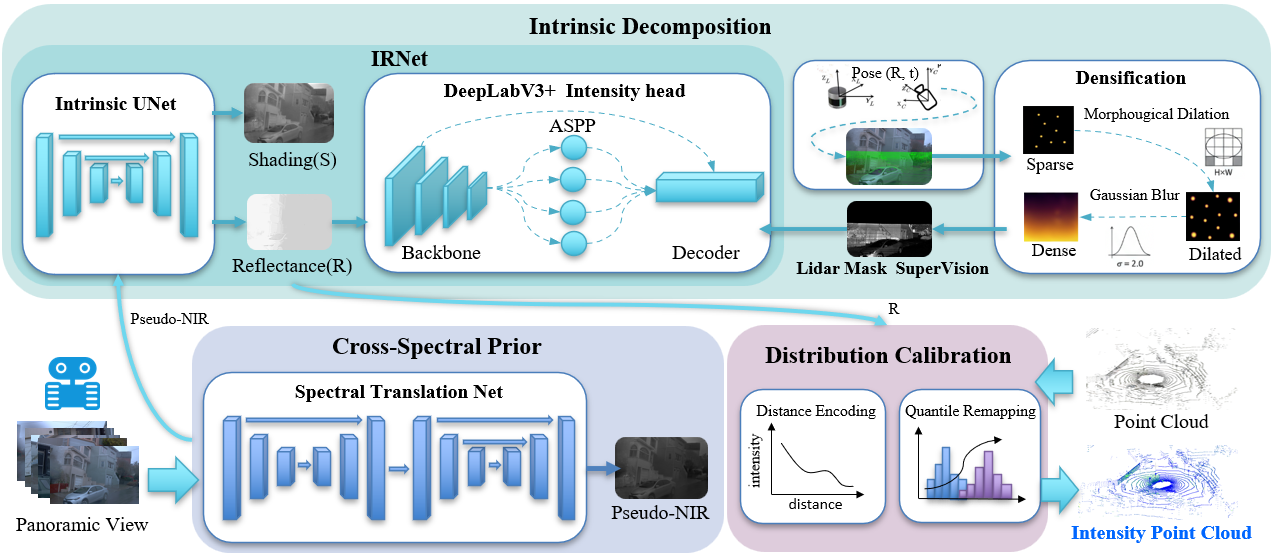}
\caption{\textbf{Overview of the NIDAR pipeline.} The framework first synthesizes pseudo-NIR from RGB inputs using the Spectral Translation Network (STN). IRNet then predicts reflectance-related ($R$), shading ($S$), and dense intensity representations under sparse projected-LiDAR supervision. Finally, calibrated projection, geometry-aware modulation, and quantile remapping assign an intensity value to each visible 3D point.}
\label{fig:pipeline8}
\vspace{-3mm}
\end{figure*}

\section{Methodology}
\vspace{-1mm}
\subsection{Problem Formulation} 
We study LiDAR intensity synthesis given RGB imagery and 3D geometry. The input comprises one or more RGB images with known camera intrinsics and extrinsics and a 3D point cloud (from a simulator or reconstruction) with camera--LiDAR calibration. The output is a point cloud augmented with an intensity-like value per visible point. NIDAR proceeds by (i) synthesizing pseudo-NIR appearance from RGB, (ii) decomposing it into reflectance-related and shading components, and (iii) transferring a learned intensity representation to 3D points via calibrated projection and geometry-aware modulation. Under the standard intrinsic image model, the pseudo-NIR observation satisfies
\begin{equation}
I \approx R \cdot S,
\label{eq:intrinsic_model}
\end{equation}
where $I$ is pseudo-NIR appearance, $R$ is a reflectance-related latent component, and $S$ is shading. Importantly, $R$ is not calibrated LiDAR return intensity. A measured return can be abstracted as
\begin{equation*}
I_{\mathrm{LiDAR}}=g_s\,\rho(m,\lambda,\theta)\,a(d,\theta)+\eta,
\end{equation*}
where $g_s$ is sensor gain, $\rho$ depends on material, wavelength, and incidence angle, $a$ collects geometric attenuation, and $\eta$ is measurement noise. NIDAR uses $R$ only as a cue; geometry-aware modulation and calibration account for part of the remaining variation, without claiming sensor-independent reconstruction of absolute returned power.

\subsection{NIDAR Framework Overview} 
As shown in Fig.~\ref{fig:pipeline8}, NIDAR comprises three stages: (i) cross-spectral pseudo-NIR synthesis from RGB using a pretrained STN; (ii) intrinsic decomposition and dense intensity prediction with IRNet under sparse LiDAR-projection supervision; and (iii) calibrated projection, geometry-aware modulation, and quantile remapping. The learned components and source-domain calibration are obtained offline. During evaluation, network weights remain fixed and no scene-specific gradient optimization is performed. This design targets scalable reuse across the evaluated scenes, while sensor-profile changes may still require calibration or fine-tuning.

\subsection{Cross-spectral Pseudo-NIR Prior}
LiDAR measures returns in the NIR band, whereas cameras capture visible-spectrum appearance. To reduce this spectral gap, we synthesize a pseudo-NIR image from RGB using the pretrained Spectral Translation Network (STN) from cs-stereo~\cite{cs-stereo2019}. Given an RGB image $I^{rgb}$, the STN outputs a single-channel pseudo-NIR image:
\begin{equation}
I^{nir} = f_{\mathrm{stn}}(I^{rgb}).
\label{eq:pseudo_nir}
\end{equation}
This stage is inference-only in NIDAR, and thus we directly use the pretrained cs-stereo STN without fine-tuning.

Qualitative results on real datasets are deferred to Sec.~\ref{sec:experiments}.

\subsection{IRNet: Intrinsic Reflectance Network}
\textbf{Intrinsic Decomposition.} We estimate reflectance $R$ and shading $S$ from pseudo-NIR input using a UNet-style encoder--decoder~\cite{ronneberger2015unet}. The decomposition is constrained by the intrinsic image formation model:
\begin{equation}
I_{\mathrm{recon}} = R \odot S, \qquad
\mathcal{L}_{\mathrm{recon}} = \lVert I_{\mathrm{recon}} - I^{nir}\rVert_1,
\label{eq:recon}
\end{equation}
where $I_{\mathrm{recon}}$ is the reconstruction and $\odot$ denotes element-wise multiplication.
To encode intrinsic priors in a fully differentiable form, we impose edge-aware smoothness on reflectance and a low-frequency prior on shading. Let $\nabla_x(\cdot)$ and $\nabla_y(\cdot)$ denote finite differences. Then,
\begin{equation}
w_x = \exp\!\big(-\gamma\,|\nabla_x I^{nir}|\big), \quad
w_y = \exp\!\big(-\gamma\,|\nabla_y I^{nir}|\big),
\label{eq:edge_weights}
\end{equation}
\begin{equation}
\mathcal{L}_{R} =
\mathbb{E}\!\left[w_x\,|\nabla_x R| + w_y\,|\nabla_y R|\right], \qquad
\mathcal{L}_{S} = \lVert S - \mathrm{AvgPool}(S)\rVert_1,
\label{eq:smooth}
\end{equation}
where $\nabla_x, \nabla_y$ are spatial gradients, $w_x,w_y$ are edge-aware weights computed from $I^{nir}$, and $\mathrm{AvgPool}(\cdot)$ is a local averaging operator that favors slowly varying shading. We further stabilize the global scale of $R$ using a mean regularizer:
\begin{equation}
\mathcal{L}_{\mathrm{med}} = \left|\mathrm{mean}(R) - m_0\right|,
\label{eq:median}
\end{equation}
where $m_0$ is the target reflectance mean (set to $0.5$ in our experiments).

\textbf{Intensity Supervision with Masks.} Intrinsic decomposition alone does not ensure agreement with measured LiDAR intensity. We therefore attach a DeepLabV3-style head~\cite{chen2017deeplabv3} that predicts dense intensity $\hat I$ under sparse projected-LiDAR supervision $I^{\mathrm{LiDAR}}$. Because the camera and LiDAR fields of view overlap only partially, $M$ masks valid projected pixels. Let $\mathcal S=\{1,\frac12,\frac14\}$, $D_s$ denote bilinear downsampling, and $M_s$ the nearest-neighbor downsampled mask. Define $E=\hat I-I^{\mathrm{LiDAR}}$, $E_s=D_s(\hat I)-D_s(I^{\mathrm{LiDAR}})$, and $S_I=\mathrm{SSIM}_{\mathrm{map}}(\hat I,I^{\mathrm{LiDAR}})$. We use
\begin{equation}
\begin{aligned}
\mathcal L_{1}^{\mathrm{ms}}&=\frac{1}{|\mathcal S|}\sum_{s\in\mathcal S}
\frac{\|M_s\odot E_s\|_1}{\|M_s\|_1+\epsilon},\\
\mathcal L_2&=\frac{\|M\odot E\|_2^2}{\|M\|_1+\epsilon},\\
\mathcal L_{\mathrm{SSIM}}&=1-\frac{\sum M\odot S_I}{\|M\|_1+\epsilon},\\
\mathcal L_{\mathrm{int}}&=w_1\mathcal L_{1}^{\mathrm{ms}}+w_2\mathcal L_2+w_{\mathrm{ssim}}\mathcal L_{\mathrm{SSIM}}.
\end{aligned}
\label{eq:LiDAR_mask}
\end{equation}
Here, reductions are spatial; binary-mask $\ell_1$ norms count valid pixels, and $\epsilon>0$ prevents division by zero.

\textbf{Overall Objective.} The complete training objective used for the reported model is
\begin{equation}
\begin{aligned}
\mathcal{L} =\ &\lambda_{\mathrm{recon}}\mathcal{L}_{\mathrm{recon}}
+ \lambda_{R}\mathcal{L}_{R}
+ \lambda_{S}\mathcal{L}_{S} \\
&+ \lambda_{\mathrm{med}}\mathcal{L}_{\mathrm{med}}
+ \lambda_{\mathrm{int}}\mathcal{L}_{\mathrm{int}}.
\end{aligned}
\label{eq:total_loss}
\end{equation}
The intensity weight is linearly warmed up during the initial training epochs. The reported model does not use a VGG perceptual loss. For comparison, we also consider IIW-CRF~\cite{bell2014intrinsic}, a classical dense-CRF intrinsic decomposition method that directly predicts $R$ and $S$ from $I^{nir}$.

\subsection{Distribution Calibration via Quantile Remapping}
The reflectance-related output can exhibit global statistical mismatch relative to real LiDAR measurements. We calibrate this distribution in two steps: a distance-dependent linear modulation followed by a learned quantile mapping. In the main pipeline, this calibration is applied to the reflectance-like $R$ branch before point assignment; direct dense-intensity predictions $\hat I$ are reported without this remapping unless explicitly marked as a post-hoc diagnostic.

\textbf{Distance Encoding.} We first apply distance-dependent linear modulation to per-point intensity to capture first-order range attenuation. Let $d = \|\mathbf{x}\|$ be the point distance and $t = (d - d_{\min})/(d_{\max} - d_{\min})$ the normalized distance in $[0,1]$ over a fixed range $[d_{\min}, d_{\max}]$. We set
\begin{equation}
i_d = R_{\max} + (R_{\min} - R_{\max})\, t, \quad \text{with } R_{\max} = i,
\label{eq:dist_encode}
\end{equation}
where $i$ is the branch value sampled at the point. Based on this value, each return is categorized as diffuse or retro-reflective: if $i \le \tau$, we treat it as diffuse and set $R_{\min}=0$; otherwise, we set $R_{\min}=r_{\mathrm{retro}}$. We then clip $i_d$ to $[0,1]$ and pass it to the quantile mapping stage.

\textbf{Quantile Remapping.} Given distance-modulated intensity $i_d$ and a reference distribution $\{r_n\}$ from source-domain LiDAR training data, we compute quantiles at $\{q_k\}_{k=1}^{K}$ to obtain source and reference curves $Q_s$, $Q_r$, and define a piecewise-linear mapping:
\begin{equation}
g(i) = \mathrm{interp}\big(i, Q_s, Q_r\big), \quad
i' = \mathrm{clip}(g(i),0,1),
\label{eq:quantile}
\end{equation}
where $Q_s$ and $Q_r$ are source and reference quantile curves evaluated at fixed levels $\{q_k\}_{k=1}^{K}$, and $\mathrm{interp}(\cdot,\cdot,\cdot)$ denotes piecewise-linear interpolation. The mapping is learned offline and held fixed during scene evaluation; inference applies Eq.~\ref{eq:quantile} without scene-specific gradient optimization. Remapping is performed only on valid points, while unobserved points remain zero.

\subsection{Point Cloud Intensity Assignment}
Finally, we project 3D points, sample $R$ (or $\hat I$), and fuse cameras when available. Let $\mathbf p_e^h=[X_e,Y_e,Z_e,1]^\top$ be a point in homogeneous coordinates, expressed in the ego/LiDAR frame, and let $\mathbf T_{e\leftarrow c}$ be the camera-to-ego extrinsic transform. We compute
\begin{equation}
\mathbf p_c^h=\mathbf T_{c\leftarrow e}\mathbf p_e^h,
\qquad \mathbf T_{c\leftarrow e}=\mathbf T_{e\leftarrow c}^{-1}.
\label{eq:extrinsic}
\end{equation}
The first three components of $\mathbf p_c^h$ are denoted by $(X_c,Y_c,Z_c)$. In the Waymo convention, $X_c$ points forward, $Y_c$ left, and $Z_c$ upward, while image coordinates increase rightward and downward. Thus, for $X_c>0$,
\begin{equation}
\begin{aligned}
x_n&=-\frac{Y_c}{X_c}, & y_n&=-\frac{Z_c}{X_c},\\
u&=f_xx_n+c_x, & v&=f_yy_n+c_y.
\end{aligned}
\label{eq:proj}
\end{equation}
Here, $(f_x,f_y,c_x,c_y)$ are the camera intrinsics. The negative signs convert the left/up camera axes to the right/down image axes; thus, Eq.~\ref{eq:proj} is a pinhole projection expressed in the Waymo coordinate convention. Dataset-specific extrinsics are applied before this projection, and no additional learnable component is introduced.

\begin{table*}[t!]
    \small
    \centering
    \caption{Quantitative results on nuScenes with reconstruction methods.}
    \renewcommand{\arraystretch}{1.2} 
    \setlength{\tabcolsep}{2.5pt} 
    \label{tab:nuScenes_exp}
    
    \resizebox{0.85\textwidth}{!}{%
    \begin{tabular}{l ccccccc c ccccc}
        \hline
        \multirow{2}{*}{{Method/Scenario}}
        & \multirow{2}{*}{{Training Paradigm}}
        & \multicolumn{5}{c}{{scene0001-frame00006}}
        &
        & \multicolumn{5}{c}{{scene0004-frame00016}}
        \\
        \cline{3-7}  \cline{9-13}
        & & RMSE↓ & MedAE↓ & LPIPS↓ & SSIM↑ & PSNR↑ & 
        & RMSE↓ & MedAE↓ & LPIPS↓ & SSIM↑ & PSNR↑ \\
        \hline
        LiDAR-NeRF
        & scene-specific & 7.6498 & 28.0431 & \nd{0.3713} & 0.5861 & 30.3280 &
        & 7.7252 & 28.0275 & 0.3838 & 0.6186 & 30.3726\\
        LiDAR4D 
        & scene-specific & 8.1914 & 32.0000 & 0.4058 & 0.6317 & 29.8636 &
        & 7.8921 & 28.0000 & 0.3695 & 0.6675 & 30.1869\\
        LiDAR-RT 
        & scene-specific & 7.9996 & 30.0430 & 0.4569 & 0.6264 & 30.0694 &
        & 7.8641 & 28.0217 & 0.3915 & 0.6371 & 30.2178\\
        NIDAR w/o. IRNet 
        & pretrained(zero shot) & \nd{7.7274} & \nd{25.0392} & 0.3932 & \nd{0.6661} & \nd{30.3701} &
        & \nd{7.5271} & \nd{27.0314} & \nd{0.3385} & \nd{0.7098} & \nd{30.5982}\\
        \textbf{NIDAR-Full (Ours)} 
        & pretrained(zero shot) & \fs\bf{7.4855} & \fs\bf{22.0392} & \fs\bf{0.3211} & \fs\bf{0.7166} & \fs\bf{30.6464} &
        & \fs\bf{7.4068} & \fs\bf{21.0431} & \bf\fs{0.2984} & \bf\fs{0.7370} & \bf\fs{30.7382}\\
        \hline
    \end{tabular}%
    }
    \vspace{-3mm}
\end{table*}

\begin{table*}[t!]
    \small
    \centering
    \caption{Quantitative results on Waymo with reconstruction methods.}
    \renewcommand{\arraystretch}{1.2} 
    \setlength{\tabcolsep}{2.5pt} 
    \label{tab:Waymo_exp}
    
    \resizebox{0.85\textwidth}{!}{%
    \begin{tabular}{l ccccccc c cccccc} 
        \hline
        \multirow{2}{*}{{Method/Scenario}}
        & \multirow{2}{*}{{Training Paradigm}}
        & \multicolumn{5}{c}{{WS4-frame000000}}
        &
        & \multicolumn{5}{c}{{WS4-frame000184}}
        \\
        \cline{3-7}  \cline{9-13} 
        & & RMSE↓ & MedAE↓ & LPIPS↓ & SSIM↑ & PSNR↑ & 
        & RMSE↓ & MedAE↓ & LPIPS↓ & SSIM↑ & PSNR↑ \\
        \hline
         LiDAR-NeRF
        & scene-specific & {0.1124} & {0.0440} & {0.6115} & {0.2813} & {18.5936} & 
        & 0.1364 & 0.0524 & 0.5987 & 0.2574 & 17.2539 \\
         LiDAR4D 
        & scene-specific & \fs\bf{0.0981} & \fs\bf{0.0308} & \nd{0.3104} & 0.5539 & \fs\bf{20.1773} & 
        & \nd{0.1144} & \fs\bf{0.0374} & 0.3613 & 0.4933 & \nd{18.8280} \\
        LiDAR-RT 
        & scene-specific & {0.1203} & {0.0573} & {0.6211} & {0.2943} & {18.3913} & 
        & {0.1288} & {0.0608} & {0.6152} & {0.2860} & {17.7984} \\
        NIDAR w/o. IRNet 
        & pretrained(zero shot) & 0.2571 & 0.1058 & 0.3464 & \nd{0.5556} & 11.7973 & 
        & 0.1588 & 0.0862 &\nd{0.2984} & \nd{0.5796} & 15.9845 \\
        \textbf{NIDAR-Full (Ours)} 
        & pretrained(zero shot) & \nd{0.1092} & \nd{0.0568} & \fs\bf{0.2151} & \fs\bf{0.6826} & \nd{19.2321} & 
        & \fs\bf{0.1132} & \nd{0.0431} & \fs\bf{0.2757} & \fs\bf{0.6891} & \fs\bf{18.9212} \\
        \hline
    \end{tabular}%
    }
    \vspace{-3mm}
\end{table*}

\section{Experiments and Evaluations}
\label{sec:experiments}
\subsection{Experimental Setup}

\textbf{Implementation Details.}
IRNet and the source-domain quantile-remapping model are trained using 17 Waymo training scenarios on a single RTX 3090 GPU. IRNet takes the pseudo-NIR images from Eq.~\ref{eq:pseudo_nir} as input and is supervised by projected LiDAR intensity at valid pixels. We train for 10 epochs with a batch size of 8 using Adam, a learning rate of $10^{-4}$, and a weight decay of $10^{-5}$. We set $(\lambda_{\mathrm{recon}},\lambda_R,\lambda_S,\lambda_{\mathrm{med}},\lambda_{\mathrm{int}})=(1.0,0.01,0,0.01,0.5)$ and $(w_1,w_2,w_{\mathrm{ssim}})=(1.0,0.5,0.2)$. Multi-scale $\ell_1$ supervision uses scales $\{1,\frac{1}{2},\frac{1}{4}\}$, and the VGG perceptual loss is disabled. The pseudo-NIR stage uses the pretrained cs-stereo STN~\cite{cs-stereo2019} without fine-tuning. All learned network weights and the source-domain calibration are held fixed during scene evaluation.

\textbf{Datasets and Baselines.}
We evaluate NIDAR on sequences from two public benchmarks with LiDAR intensity ground truth: nuScenes~\cite{caesar2020nuscenes} and Waymo~\cite{sun2020waymo}. On Waymo, we use only points from the roof-mounted top LiDAR that fall within available camera views; the four side LiDARs are not evaluated. We compare against LiDAR-NeRF~\cite{tao2023lidarnerf}, LiDAR4D~\cite{zheng2024lidar4d}, and LiDAR-RT~\cite{2025lidarrt}. LiDAR-NeRF is NeRF-based, whereas LiDAR4D and LiDAR-RT adopt 3D Gaussian splatting~\cite{kerbl2023gaussian}. We use their official implementations and reported hyperparameters. Scene-specific baselines are optimized using 20 consecutive frames on nuScenes and 50 consecutive point clouds on Waymo; NIDAR performs no gradient-based fitting on the evaluated scenes.

\textbf{Metrics.}
Following prior work~\cite{wang2004ssim,zhang2018lpips}, we report MedAE, RMSE~\cite{chai2014rmse}, PSNR~\cite{huynh2008psnr}, SSIM~\cite{wang2004ssim}, and LPIPS~\cite{zhang2018lpips} between predicted and ground-truth LiDAR intensity projection maps.

\textbf{Controlled Diagnostic Protocol.}
To isolate the role of the pseudo-NIR prior from architecture and data-size confounds, we additionally train two matched IRNet variants with identical architecture, loss weights, and optimization settings: one receives STN-generated pseudo-NIR input, and the other receives RGB input directly. Both variants use the full intrinsic objective, train on the same RGB/pseudo-NIR intersection from one Waymo-style training scene, and are evaluated on held-out frames from two scenes. We report the direct dense-intensity head $\hat I$ and the reflectance-related output $R$ before and after quantile remapping. Remapping is applied only to $R$, because it was designed for reflectance-like outputs rather than for the direct $\hat I$ branch.

\subsection{Results on Public Benchmarks}
\vspace{-1mm}

\begin{figure*}[htbp]
\centering
\includegraphics[width=0.9\linewidth]{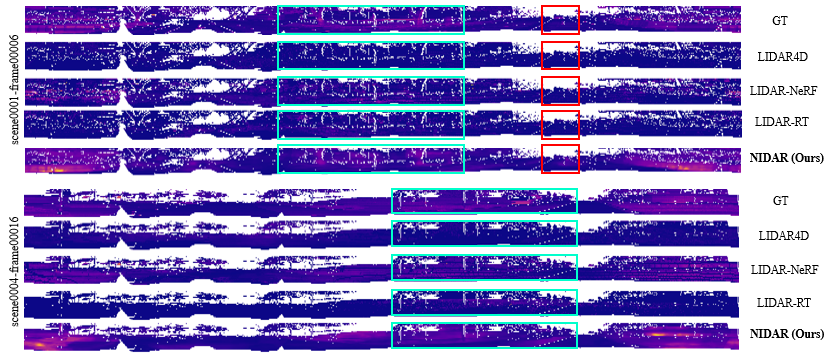}
\caption{\textbf{Qualitative comparison on nuScenes with reconstruction methods.} Colored boxes highlight corresponding regions for close comparison; the same color denotes the same spatial location across methods.}
\label{fig:nuScenes_exp}
\vspace{-3mm}
\end{figure*}

Reconstruction-based baselines require per-scene training data, whereas NIDAR runs directly on the chosen evaluation frames with fixed network weights. For Waymo, we exclude rear-view point clouds because corresponding camera images are unavailable. Tables~\ref{tab:nuScenes_exp} and~\ref{tab:Waymo_exp} summarize the quantitative results, and Fig.~\ref{fig:nuScenes_exp} provides qualitative comparisons.

On nuScenes, where each evaluated scenario contains 20 frames, reconstruction-based baselines have a short scene-fitting window and can produce over-smoothed or locally unstable intensity patterns (Fig.~\ref{fig:nuScenes_exp}). NIDAR-Full shows clearer object-aligned transitions and more coherent local contrast in the reported examples. These observations are consistent with the intended role of intrinsic decomposition and dense intensity supervision, although the current experiments do not fully isolate every source of improvement or cascading error.

On Waymo, the longer fitting window makes reconstruction baselines competitive in absolute error, while NIDAR-Full remains competitive in pixel-wise fidelity and obtains favorable SSIM and LPIPS. Although Waymo records five LiDARs, we evaluate only its roof-mounted top LiDAR; nuScenes uses a single 32-beam Velodyne HDL-32E. Differences in beam patterns, return representations, sampling density, camera pipelines, and environments may affect intensity distributions and metrics. Because our experiments do not isolate these factors, we compare methods within each dataset and avoid interpreting cross-dataset gaps as sensor-independent radiometric performance.


\subsection{Ablation Studies}
\label{sec:ablation}
\vspace{-1mm}
We analyze the design choices of the proposed IRNet. We first ablate the intrinsic backbone and the intensity head (Tab.~\ref{tab:ablation}). We then compare against the classical IIW-CRF baseline on the MIT Intrinsic Images dataset~\cite{grosse2009intrinsic} (Tab.~\ref{tab:Compare_ablation} and Fig.~\ref{Compare_ablation}).

\begin{table}[t!]
    \centering
    \caption{IRNet structural ablation.}
    \label{tab:ablation}
    \renewcommand{\arraystretch}{1.2}
    \setlength{\tabcolsep}{3pt}
    \resizebox{0.9\columnwidth}{!}{%
    \begin{tabular}{l c c c c}
        \hline
        \multirow{2}{*}{\textbf{Method}} & \multicolumn{4}{c}{\textbf{Average Performance}} \\
        \cline{2-5} 
         & RMSE↓ & MAE↓ & SSIM↑ & PSNR↑ \\
        \hline
        U-Net & {0.2306} & 0.1873 & 0.2370 & 12.8903\\
        Deeplabv3 & \rd{0.2266} & {0.1854} & 0.2501 & 13.0103\\
        U-Net+3Conv & {0.2291} & {0.1770} & 0.2624 & 12.9518\\
        U-Net+4Conv & 0.2390 & {0.1764} & 0.2677 & 12.5610\\
        U-Net+U-Net & 0.2500 & 0.2095 & 0.2532 & 12.2134\\
        Deeplabv3+3Conv & \nd{0.2252} & \rd{0.1735} & 0.2880 & \rd{13.3529}\\
        Deeplabv3+4Conv & 0.2365 & {0.1803} & 0.2870 & 13.0072\\
        Deeplabv3+U-Net & {0.2318} & {0.1754} & \rd{0.2967} & 13.3274\\
        Deeplabv3+Deeplabv3 & {0.2279} & \nd{0.1723} & \nd{0.2972} & \nd{13.3960}\\
        \textbf{Ours (U-Net+Deeplabv3)} & \fs\bf{0.2242} & \fs\bf{0.1666} & \fs\bf{0.3209} & \fs\bf{13.6547}\\
        \hline
    \end{tabular}%
    }
    \vspace{-3mm}
\end{table}

To determine the optimal architecture, we conduct a combinatorial ablation over the intrinsic backbone (UNet, DeepLabV3), the intensity head (Conv, UNet, DeepLabV3), and the head depth when using Conv. The Conv head denotes stacked $3\times3$ convolutions with BatchNorm and ELU. Accordingly, ``3Conv'' and ``4Conv'' indicate 3- and 4-layer stacks. Each configuration is trained for 10 epochs on 6 Waymo scenes~\cite{sun2020waymo} with identical loss weights. We evaluate using RMSE and MAE~\cite{chai2014rmse}, SSIM~\cite{wang2004ssim}, and PSNR~\cite{huynh2008psnr} on three held-out test scenes. As shown in Tab.~\ref{tab:ablation}, the hierarchical configuration U-Net+DeepLabV3 achieves the best overall performance.

Table~\ref{tab:p0_branch_control} reports a controlled diagnostic of where the pseudo-NIR prior helps. The RGB and pseudo-NIR variants use the same architecture, intrinsic objective, optimization settings, and frame-camera pairs: the exact RGB/pseudo-NIR intersection from one Waymo-style training scene (495 pairs, split into 445 training and 50 validation pairs). We evaluate both variants on 20 held-out frames from each of two scenes. The sparse intensity validation losses are nearly identical, with RGB slightly lower than pseudo-NIR ($0.02044$ vs. $0.02049$), so the comparison is not explained by one input being easier to fit under the direct LiDAR supervision.

The benefit of pseudo-NIR is branch-dependent. For the direct dense-intensity head $\hat I$, RGB gives lower MAE/RMSE, while pseudo-NIR gives slightly higher SSIM. This indicates that RGB remains a strong input for direct pointwise intensity regression, and pseudo-NIR should not be interpreted as a universal shortcut for the supervised head. The advantage appears in the intrinsic-decomposition route: when the output is the reflectance-related $R$, pseudo-NIR substantially improves MAE, RMSE, and SSIM before remapping, and after source-domain remapping it produces the strongest structural score while retaining lower MAE/RMSE than the RGB $R$ route. In representation-learning terms, pseudo-NIR provides an inductive bias that makes the intermediate $R$ representation more structurally aligned and more amenable to calibration, which matches the intrinsic-decomposition hypothesis used by NIDAR.

\begin{table}[t]
    \small
    \centering
    \caption{Pseudo-NIR intermediate-state ablation.}
    \label{tab:p0_branch_control}
    \setlength{\tabcolsep}{2.5pt}
    \renewcommand{\arraystretch}{1.1}
    \resizebox{\columnwidth}{!}{%
    \begin{tabular}{l c c c c c c}
        \toprule
        \textbf{Input} & \makecell{\textbf{Val.}\\\textbf{loss}} & \textbf{Output} & \textbf{Post.} & \textbf{MAE$\downarrow$} & \textbf{RMSE$\downarrow$} & \textbf{SSIM$\uparrow$} \\
        \midrule
        \multirow{3}{*}{\scriptsize RGB} & \multirow{3}{*}{0.02044} & $\hat I$ & Raw & \fs\bf{0.1611} & \fs\bf{0.1914} & 0.3772 \\
        & & $R$ & Raw & 0.2935 & 0.3330 & 0.2564 \\
        & & $R$ & Remap & \rd{0.1764} & 0.2244 & 0.3698 \\
        \midrule
        \multirow{3}{*}{\scriptsize Pseudo-NIR} & \multirow{3}{*}{0.02049} & $\hat I$ & Raw & 0.1847 & \rd{0.2187} & \nd{0.3826} \\
        & & $R$ & Raw & 0.1883 & 0.2436 & \rd{0.3792} \\
        & & $R$ & Remap & \nd{0.1621} & \nd{0.2105} & \fs\bf{0.4032} \\
        \bottomrule
    \end{tabular}%
    }
    \vspace{-2mm}
\end{table}

\begin{table}[!htbp]
    \caption{IRNet and IIW-CRF on MIT Intrinsic dataset~\cite{grosse2009intrinsic}.}
    \centering
    \setlength{\tabcolsep}{2.5pt}
    \resizebox{\columnwidth}{!}{%
    \begin{tabular}{l c c c c c c}
        \hline
        \textbf{Methods} & \makecell{T-Infer$^1$\\(s/frame)} & RMSE↓ & MedAE↓ & LPIPS↓ & SSIM↑ & PSNR↑ \\ 
        \hline
        IIW-CRF~\cite{bell2014intrinsic} & 100.72 & 0.1465 & 0.1054 & 0.5491 & \fs \bf0.6897 & 16.6846\\ 
        IRNet (Ours) & \fs \bf{0.0438} & \fs \bf{0.1394} & \fs \bf0.1019 & \fs \bf0.3684 & 0.6747 & \fs \bf17.1117\\ 
        \hline
    \end{tabular}%
    }
    \label{tab:Compare_ablation}
    \begin{flushleft}
        \scriptsize{$^1$ T-Infer measures the intrinsic-decomposition stage only; it excludes STN, calibration, projection, and I/O.}
    \end{flushleft}
    \vspace{-3mm}
\end{table}

\begin{figure}[!htbp]
    \centering
    \setlength{\tabcolsep}{1pt}

    \includegraphics[width=1.00\linewidth]{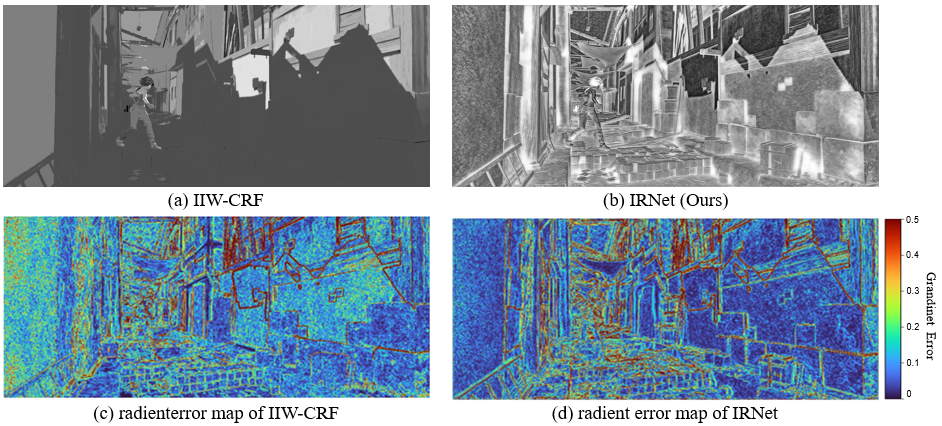}
    \vspace{-3mm}

    \caption{\textbf{IRNet and IIW-CRF on MIT Intrinsic dataset.} Gradient error is defined as the absolute difference between the predicted gradient magnitude and the ground-truth gradient magnitude, where lower values indicate better texture and edge reconstruction.}
    \label{Compare_ablation}
    \vspace{-3mm}
\end{figure}

\begin{figure}[htbp]
    \centering
    \includegraphics[width=0.95\columnwidth]{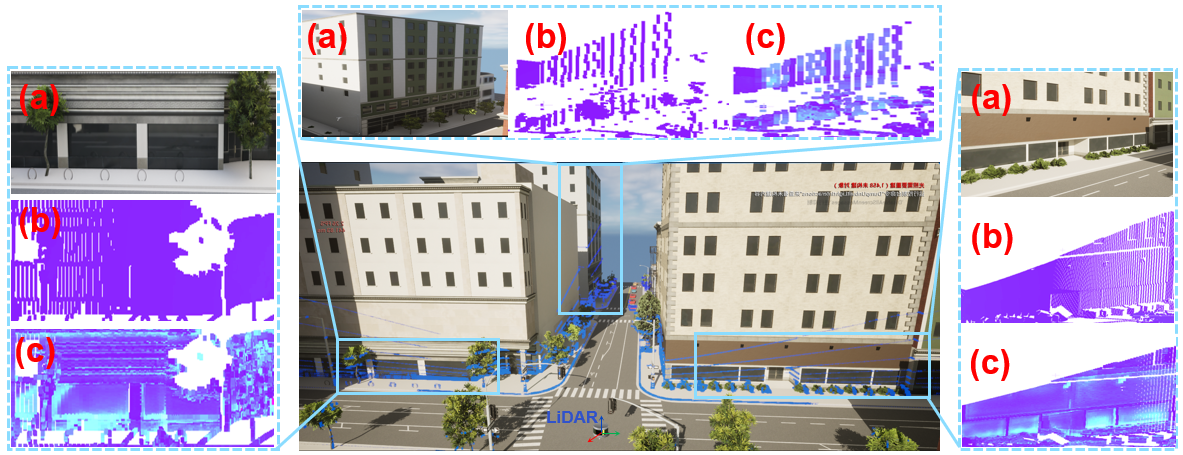}
    \caption{\textbf{Integration of NIDAR with Unreal Engine 5.} (a) Rendered scene. (b) Raw simulated LiDAR point cloud. (c) Point cloud augmented with NIDAR intensity. Colors visualize normalized intensity magnitude and do not denote semantic classes.}
    \label{fig:ue5}
    \vspace{-3mm}
\end{figure}

\subsection{Applications}

\begin{figure}[t]
\centering
  \includegraphics[width=0.4\textwidth]{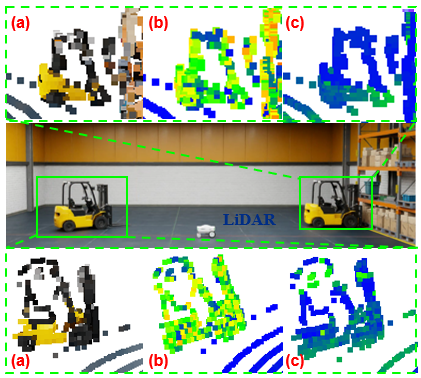}
  \caption{\textbf{NIDAR integration in Isaac Sim.} (a) RGB-colored point cloud. (b) Built-in simulated intensity. (c) NIDAR intensity. In the shown example, NIDAR produces smoother and more spatially coherent intensity patterns. Colors visualize normalized intensity magnitude and do not denote semantic classes.}
  \label{isaacsim}
\vspace{-0.3cm}
\end{figure}

\textbf{Integration with Graphics Engines.}
We integrate NIDAR with UE5 and Isaac Sim, as shown in Figs.~\ref{fig:ue5} and~\ref{isaacsim}. The examples demonstrate that the interface can augment simulator point clouds with spatially coherent intensity patterns. Because matched real-LiDAR ground truth is unavailable for these rendered scenes, the figures are qualitative demonstrations of pipeline compatibility rather than quantitative evidence of absolute physical or sensor fidelity.

\begin{figure}[t] 
    \centering
    \begin{minipage}{\columnwidth}
        \centering
        \includegraphics[width=0.85\columnwidth]{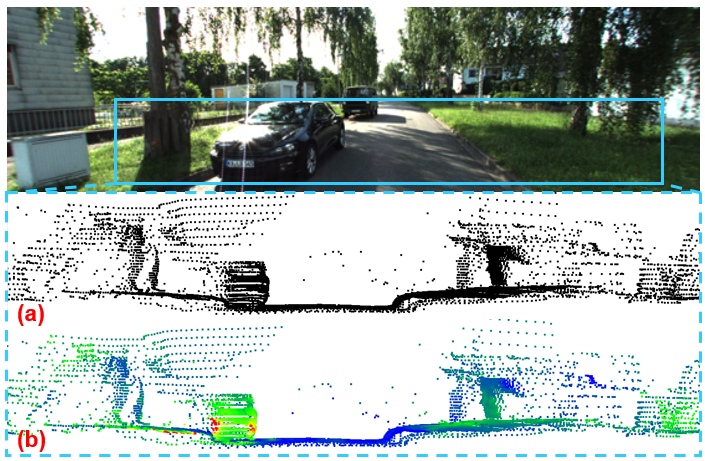}
    \end{minipage}
       
    \vspace{-2mm}
    \caption{\textbf{NIDAR integration with LiDAR-Diffusion~\cite{ran2024towards}.} (a) Point cloud generated by LiDAR-Diffusion. (b) The same point cloud augmented with NIDAR intensity.}
    \label{fig:LiDARdiffusion}
    \vspace{-4mm}
\end{figure}

\textbf{Integration with Generative Methods.}
For generative LiDAR simulation, NIDAR post-processes image-conditioned LiDAR-Diffusion outputs~\cite{ran2024towards}, assigning RGB-derived intensity to camera-visible generated points through calibrated projection. This preserves geometry without per-scene NIDAR optimization (Fig.~\ref{fig:LiDARdiffusion}).

    
    
    

\subsection{Downstream SLAM Evaluation}

To examine the potential downstream utility of NIDAR-generated intensity,
we conduct a simulated SLAM case study.
We collect trajectories with LiDAR measurements from two indoor scenes
({Warehouse} and {Hospital}) in Isaac Sim, 
and evaluate two representative intensity-aware LiDAR SLAM systems:
Intensity-SLAM~\cite{wang2021intensity} 
and ISC-LOAM~\cite{ISCLOAMwang2020intensity}.
For each method, we compare SLAM performance using (i) no intensity input (when applicable), 
(ii) the built-in simulated intensity provided by Isaac Sim, 
and (iii) the intensity inferred by NIDAR, 
while keeping all other inputs and parameters unchanged. 
We report the Absolute Trajectory Error (ATE) in terms of RMSE, 
computed against the ground-truth trajectory provided by the simulator.

Table~\ref{tab:slam} reports the simulated ATE. Among successful runs, NIDAR has the lowest ATE for both backends in both scenes. The Hospital gain over built-in intensity is marginal for Intensity-SLAM ($0.1100$ to $0.1085$~m) but larger for ISC-LOAM ($3.2578$ to $1.9187$~m). In Warehouse, NIDAR enables Intensity-SLAM to complete and reduces ISC-LOAM ATE from $4.1218$ to $2.3504$~m. These simulated results suggest potential utility but do not establish real-robot gains.
Additional qualitative results and full video demonstrations are available on the project website: \url{https://nidar-web.github.io/}.

\begin{figure}[t] 
    \centering
    \begin{minipage}{\columnwidth}
        \centering
        \includegraphics[width=0.9\columnwidth]{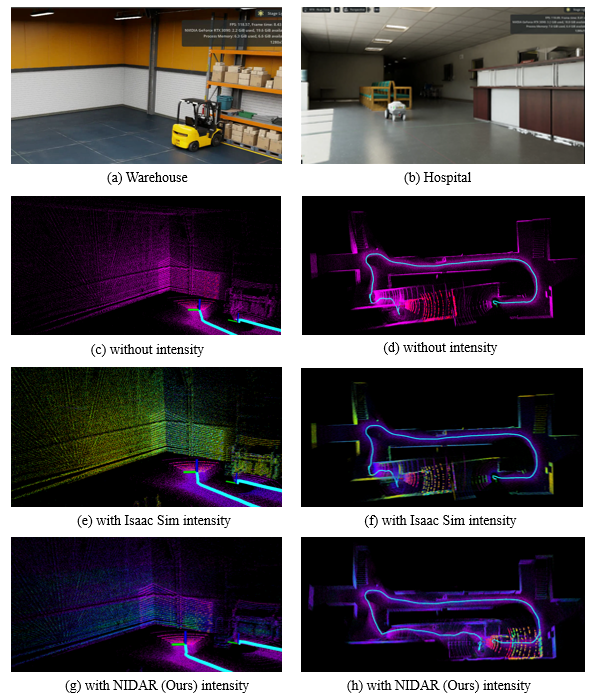}
    \end{minipage}
       
    \vspace{-2mm}
   
    \caption{\textbf{Simulated downstream SLAM case study in Isaac Sim.} The figure compares trajectories and maps obtained with no intensity, built-in intensity, and NIDAR intensity in the reported Warehouse and Hospital sequences.}
    \label{fig:slam}
    \vspace{-4mm}
\end{figure}

\begin{table}[t]
\centering
\vspace{2mm}
\caption{Simulated SLAM case study: ATE RMSE in meters. A cross indicates that the corresponding run did not complete.}
\vspace{1mm}
\setlength{\tabcolsep}{1pt}
\resizebox{\columnwidth}{!}{
\begin{tabular}{l|ccc|ccc}
\toprule
\multirow{2}{*}{Scene/Method}
& \multicolumn{3}{c|}{Intensity-SLAM\cite{wang2021intensity}}
& \multicolumn{3}{c}{ISC-LOAM\cite{ISCLOAMwang2020intensity}} \\
\cmidrule(lr){2-4} \cmidrule(lr){5-7}
& w/o intensity & w built-in intensity & w NIDAR intensity& w/o intensity & w built-in intensity & w NIDAR intensity \\
\midrule
Warehouse 
& \blackx & \blackx & \textbf{1.0570}
& 4.3988 & {4.1218} & \textbf{2.3504} \\

Hospital 
& 1.3925 & 0.1100 & \textbf{0.1085}
& 3.1070 & 3.2578 & \bf 1.9187 \\
\bottomrule
\end{tabular}%
}
\label{tab:slam}
\vspace{-3mm}
\end{table}

\section{Conclusion}
We presented NIDAR, a feed-forward framework for synthesizing dense LiDAR intensity-like observations from RGB appearance and simulator geometry. The method combines spectral translation, intrinsic decomposition, geometry-aware modulation, and distribution calibration, and requires no per-scene network optimization in the evaluated settings. Experiments on Waymo and nuScenes demonstrate competitive pixel-wise accuracy and favorable structural and perceptual fidelity in the reported comparisons, while a controlled branch diagnostic shows that pseudo-NIR is most beneficial for the reflectance-calibration route used by the main pipeline, even when direct RGB and pseudo-NIR heads fit sparse intensity supervision similarly. Integrations with three simulation pipelines show the portability of the interface, while two simulated SLAM case studies indicate potential utility for intensity-aware localization. The current evidence supports cross-scene transfer within the evaluated datasets and simulators, rather than universal sensor-independent radiometric reconstruction. Future work will focus on camera and wavelength robustness, real-sensor validation, embedded deployment, and more comprehensive downstream evaluation.


\small
\bibliographystyle{IEEEtrans}
\bibliography{root}

\end{document}